\documentclass[journal]{IEEEtran}
\ifCLASSINFOpdf
\else
\fi
\usepackage{hyperref}

\usepackage{multirow}
\usepackage{orcidlink}
\usepackage{adjustbox}
\usepackage{amssymb}
\usepackage{amsmath}
\usepackage{bbding}
\usepackage{mathtools}
\usepackage{booktabs}
\RequirePackage{fancyhdr}
\RequirePackage{xcolor} % changed from color to xcolor (2021/11/24)
\RequirePackage{algorithm}
\RequirePackage{algorithmic}
\RequirePackage{eso-pic} % used by \AddToShipoutPicture
\RequirePackage{forloop}
\RequirePackage{url}
\RequirePackage{caption}
\begin{document}
\title{Evolving, Not Training: \\Zero-Shot Reasoning Segmentation via Evolutionary Prompting}

\author{Kai Ye \orcidlink{0000-0003-0914-0213}, Xiaotong You\orcidlink{0009-0000-0378-5569}, Jianghang Lin \orcidlink{0009-0009-3873-3860}, Jiayi Ji \orcidlink{0000-0002-9956-6308}, Pingyang Dai \orcidlink{0000-0001-9780-271X}, Liujuan Cao \orcidlink{0000-0002-7645-9606}% <-this % stops a space
\thanks{ 
Kai Ye, Xiaotong You, Jianghang Lin are with the Key Laboratory of Multimedia Trusted Perception
 and Efficient Computing, Ministry of Education of China, and the Institute
 of Artificial Intelligence, Xiamen University, Xiamen 361005, China (e-mail:
 yekai@stu.xmu.edu.cn; uiosun@stu.xmu.edu.cn ; hunterjlin007@stu.xmu.edu.cn).

  Jiayi Ji is with the Key Laboratory of Multimedia Trusted Perception and
 Efficient Computing, Ministry of Education of China, Xiamen University,
 Xiamen 361005, China, and also with the National University of Singapore,
 Singapore 119077 (e-mail: jjyxmu@gmail.com).

  Liujuan Cao\textit{(Corresponding author)} and Pingyang Dai are with the Key Laboratory of Multimedia Trusted Perception and Efficient Computing, Ministry of Education
 of China, Xiamen University, Xiamen 361005, China, and also with the
 School of Informatics, Xiamen University, Xiamen 361005, China (e-mail:
 caoliujuan@xmu.edu.cn, pydai@xmu.edu.cn).
 }}

\maketitle

\begin{abstract}
Reasoning Segmentation requires models to interpret complex, context-dependent linguistic queries to achieve pixel-level localization. Current dominant approaches rely heavily on Supervised Fine-Tuning (SFT) or Reinforcement Learning (RL). However, SFT suffers from catastrophic forgetting and domain dependency, while RL is often hindered by training instability and rigid reliance on predefined reward functions. Although recent training-free methods circumvent these training burdens, they are fundamentally limited by a static inference paradigm. These methods typically rely on a single-pass "generate-then-segment" chain, which suffers from insufficient reasoning depth and lacks the capability to self-correct linguistic hallucinations or spatial misinterpretations. In this paper, we challenge these limitations and propose EVOL-SAM3, a novel zero-shot framework that reformulates reasoning segmentation as an inference-time evolutionary search process. Instead of relying on a fixed prompt, EVOL-SAM3 maintains a population of prompt hypotheses and iteratively refines them through a "Generate-Evaluate-Evolve" loop. We introduce a Visual Arena to assess prompt fitness via reference-free pairwise tournaments, and a Semantic Mutation operator to inject diversity and correct semantic errors. Furthermore, a Heterogeneous Arena module integrates geometric priors with semantic reasoning to ensure robust final selection. Extensive experiments demonstrate that EVOL-SAM3 not only substantially outperforms static baselines but also significantly surpasses fully supervised state-of-the-art methods on the challenging ReasonSeg benchmark in a zero-shot setting.
The code is available at \url{https://github.com/AHideoKuzeA/Evol-SAM3}.
\end{abstract}

% \fi
\IEEEpeerreviewmaketitle

\section{Introduction}
\IEEEPARstart{R}easoning Segmentation aims to generate pixel-level binary masks based on complex natural language queries~\cite{lisa}.
This task requires the model to interpret text instructions through logical reasoning.
Unlike traditional semantic or instance segmentation tasks that rely on pre-defined categorical labels~\cite{mask_rcnn,fcn}, reasoning segmentation addresses more complex, context-dependent queries or even instructions requiring reasoning to identify the target instance (e.g., ``something that the person uses to fish'').
This setting better reflects the real-world requirements for intelligent assistants and robotics, while simultaneously imposing significantly higher demands on the model: it must possess fine-grained visual perception, robust image-text alignment capabilities, and broad domain knowledge combined with common sense and logical reasoning.
\begin{figure}[t]
    \centering
    \includegraphics[width=0.95\linewidth]{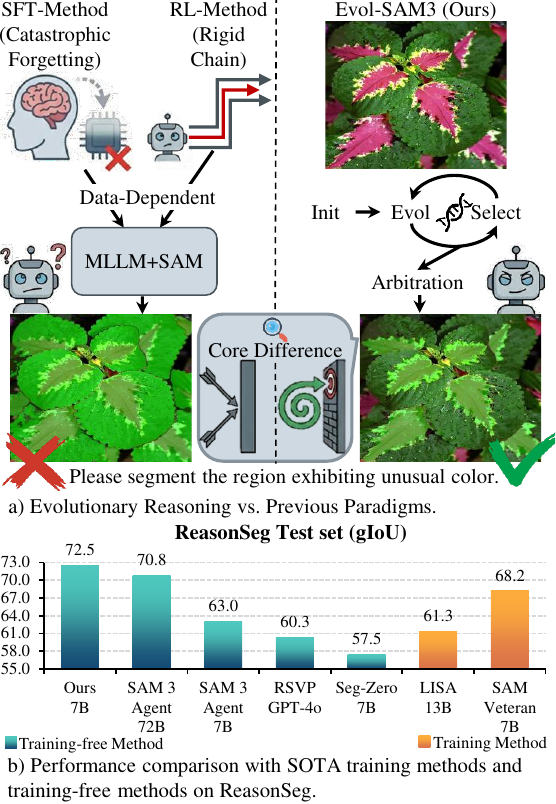}
    \caption{Core differance and performance comparison with existing methods}
    \label{fig:placeholder}
\end{figure}
Given the comprehensive capabilities of Multi-modal Large Language Models (MLLMs) in reasoning and multi-modal perception, incorporating MLLMs has become the mainstream solution for this task.

Current approaches for reasoning segmentation are primarily dominated by two learning-based paradigms: Supervised Fine-Tuning (SFT) and Reinforcement Learning (RL), yet both face inherent limitations regarding training costs and rigid inference patterns.
The first paradigm, SFT-based methods~\cite{lisa,glamm}, attempts to internalize the segmentation task by fine-tuning MLLMs to map text instructions directly to pixel masks. However, the extensive parameter updates required often induce catastrophic forgetting of general reasoning capabilities, leading to poor generalization on out-of-distribution instructions~\cite{lisa}.
The second paradigm, RL-based methods~\cite{lens,sam_veteran}, treats the MLLM as a policy network optimized via reward signals. Despite introducing dynamic interaction, these models still execute a fixed policy during inference.
This results in a brittle, linear reasoning trajectory: if the initial step falls into a local optimum (e.g., misidentifying the target), the agent lacks the intrinsic mechanism to perform dynamic backtracking or global search to recover~\cite{sam3}.

The emergence of SAM 3~\cite{sam3} marks a pivotal turning point, as it endows a powerful interactive segmentation model with intrinsic semantic understanding capability for the first time.
Unlike predecessor models that responded only to geometric prompts like points or boxes (which was the fundamental reason previous methods required expensive training for MLLMs to output coordinates~\cite{grounded_sam}), SAM 3 can directly comprehend and locate simple noun phrases.
Based on this technological leap, works represented by the SAM 3 Agent~\cite{sam3} have rapidly emerged, establishing a new paradigm of Training-Free Visual Agents. This paradigm completely discards parameter updates, utilizing a frozen SOTA MLLM as the brain for planning and SAM 3 as the hand with a semantic interface for execution.
This design maximizes the preservation of the MLLM's general reasoning abilities and world knowledge, endowing the system with strong zero-shot transfer capabilities.

However, despite the potential shown by the SAM 3 Agent, its inference mechanism remains trapped in a Heuristic Trial-and-Error strategy, suffering from two critical robustness defects.
First, the refinement steps of the SAM 3 Agent rely heavily on the MLLM's intuition to decide. This jumping logic based on dialogue history is often random and unstructured (e.g., the model may switch disorderly between color and location descriptions). Lacking a clear gradient or optimization direction, the agent easily oscillates around suboptimal solutions, failing to converge to the prompt that triggers SAM 3's peak performance.
Secondly, although the SAM 3 Agent includes a selection mechanism, it typically filters within a set of candidates generated by a single reasoning path.
This means that if the initial direction of semantic understanding deviates, all subsequent interactions are confined within this erroneous semantic frame.
The system lacks a mechanism to simultaneously maintain and compare distinct heterogeneous hypotheses, leading to an inability to escape local optima.

To address these challenges, we propose EVOL-SAM3, a novel training-free framework that reframes reasoning segmentation as an inference-time evolutionary search problem. Unlike the SAM 3 Agent, which relies on linear heuristic jumps, Evo-SAM3 maintains a dynamically evolving population of hypotheses. We design a comprehensive visual evolutionary algorithm: it conducts extensive directed exploration in the semantic space via semantic mutation operators and introduces a heterogeneous competition mechanism, forcing ``text-based reasoning'' and ``box-based localization'' to compete within the same ecological niche to resolve ambiguity. Most crucially, we leverage the MLLM's superior discriminative capability to construct a Visual Arena replacing simple filtering with a tournament selection mechanism. This enables Evo-SAM3 to transform the MLLM's discriminative power into a driving force for search without any parameter updates, ultimately evolving the optimal prompt that maximizes SAM 3's potential.

Our main contributions are summarized as follows:
\begin{itemize}
    \item We are the first to model the reasoning segmentation task as an inference-time evolutionary search problem. We demonstrate that maintaining a diverse population of hypotheses and using evolutionary algorithms for systematic search allows for a more robust approximation of the optimal solution.
    \item We introduce a specialized evolutionary mechanism featuring (1) semantic mutation for directed exploration, (2) heterogeneous competition to resolve ambiguity, and (3) a visual arena utilizing tournament selection for robust optimization.
    \item Experiments on benchmarks demonstrate that under the Zero-Shot setting, EVOL-SAM3 not only significantly outperforms existing training-free agents but also surpasses dedicated models trained via STF or RL on multiple key metrics.
\end{itemize}

\section{Related Works}
%基于学习的推理分割当前的推理分割方法大致可以分为三个范式：有监督微调（SFT）、强化学习（RL）以及混合 SFT+RL 策略。 基于 SFT 的方法~\cite{lisa} 将该任务转化为序列生成问题。它们在大规模分割数据集上端到端地微调多模态大语言模型（MLLMs），训练模型输出特定的掩码 Token 或多边形坐标。然而，这种范式面临显著的局限性。为了实现像素对齐所需的激进参数更新往往会导致灾难性遗忘，降低了 MLLM 固有的通用推理能力。此外，SFT 模型倾向于过拟合训练分布，导致在开放世界场景中遇到复杂的分布外指令时泛化能力较差。

\textbf{Learning-based Reasoning Segmentation.}
Current methodologies for reasoning segmentation can be broadly categorized into three paradigms: Supervised Fine-Tuning , Reinforcement Learning , and hybrid SFT+RL strategies.
SFT-based methods~\cite{lisa} typically formulate the task as a sequence generation problem. By end-to-end fine-tuning Multi-modal Large Language Models on large-scale segmentation datasets, these models are trained to output specific mask tokens or polygon coordinates.
However, this paradigm faces significant limitations. The aggressive parameter updates required for pixel-level alignment often induce catastrophic forgetting, degrading the MLLM's inherent general reasoning capabilities. Furthermore, SFT models tend to overfit the training distribution, resulting in poor generalization when encountering complex, out-of-distribution instructions in open-world scenarios.

%为了将优化目标与评估指标对齐，基于 RL 的方法将分割表述为策略决策过程，通过奖励信号优化模型。现有的 RL 方法通常未能实现 MLLM 与 SAM 之间完全端到端的梯度流交互，或者严重依赖人工构造的轨迹数据进行初始化 ~\cite。这种黑盒式的优化生成的提示往往是次优的，无法激发 SAM 的最佳性能。 最近的工作尝试通过 SFT+RL 策略结合两者的优势。一个代表性的方法 LENS~\cite{lens} ，通过采用包含句子级、框级和分割级奖励的组相对策略优化（GRPO），LENS 以端到端的方式联合优化了思维链和掩码生成。 尽管架构有所进步，但所有基于学习的范式——无论是 SFT、RL 还是混合策略——都有一个共同的关键瓶颈：对训练资源的过度依赖。开发此类模型需要大量的计算资源和海量标注数据集，构成了极高的准入门槛。此外，一旦训练完成，推理策略即保持静态，缺乏在不进行参数更新的情况下自适应自我纠正的灵活性。
%免训练视觉智能体针对高昂的训练门槛，一种免训练范式应运而生，以 SAM 3 Agent~\cite{sam3} 为代表。该方法将冻结的 MLLM 作为规划器与 Segment Anything Model 3 (SAM 3) 链接以进行执行，从而保留了 MLLM 固有的泛化能力。 然而，当前的免训练 Agent 严重依赖启发式的试错。它们的推理过程通常是线性的且非结构化的，缺乏系统的优化目标。如果 Agent 的初始语义解释存在缺陷，它缺乏全局探索替代假设的机制，往往导致在复杂场景中出现不可逆的失败。为了填补这一空白，我们提出了 EVOL-SAM3。与线性的启发式方法不同，我们的框架将分割重构为推理时的进化搜索过程。通过维护多样化的假设种群并利用 MLLM 的判别能力进行锦标赛选择，我们在无需训练的情况下实现了鲁棒的全局优化。
\textbf{RL and Hybrid Strategies.}
To align the optimization objective with evaluation metrics, RL-based methods formulate segmentation as a policy decision process, optimizing the model via reward signals.
However, existing RL methods often fail to achieve fully end-to-end gradient flow between the MLLM and SAM, or heavily rely on manually constructed trajectory data for initialization~\cite{seg_zero}. This black-box optimization often yields suboptimal prompts that fail to trigger the executor's peak performance.
Recent works attempt to combine the strengths of both worlds through hybrid strategies. A representative method, LENS~\cite{lens}, jointly optimizes the Chain-of-Thought (CoT) and mask generation in an end-to-end manner by employing Group Relative Policy Optimization (GRPO) with rewards spanning sentence, box, and segment levels.
Despite these architectural advancements, all learning-based paradigms—whether SFT, RL, or hybrid—share a critical bottleneck: \textbf{excessive reliance on training resources}. Developing such models requires substantial computational resources and massive annotated datasets, creating an extremely high barrier to entry. Moreover, once trained, the inference policy remains static, lacking the flexibility to adaptively self-correct without further parameter updates.

\textbf{Training-Free Visual Agents.}
In response to the high training barriers, a Training-Free paradigm has emerged, represented by the SAM 3 Agent~\cite{sam3}. This approach links a frozen MLLM as a planner with SAM 3 for execution, thereby preserving the MLLM's inherent generalization capabilities.
Nevertheless, current training-free agents rely heavily on heuristic trial-and-error. Their reasoning process is typically linear and unstructured, lacking a systematic optimization objective. If the agent's initial semantic interpretation is flawed, it lacks the mechanism to globally explore alternative hypotheses, often leading to irreversible failures in complex scenes.To address these limitations, we propose \textbf{EVOL-SAM3}, which reformulates reasoning segmentation as an \textit{inference-time evolutionary search} problem. Unlike linear approaches, our framework maintains a population of diverse hypotheses and leverages the discriminative power of MLLMs to perform global optimization, enabling robust alignment without any parameter updates.

\begin{figure*}[t]
    \centering
    \includegraphics[width=1\linewidth]{pipline.pdf}
    \caption{The overall framework of EVOL-SAM3. The pipeline transforms static reasoning segmentation into a dynamic evolutionary search process, consisting of three phases: (1) Initialization: An MLLM acts as a Meta-Generator to expand the initial query into a diverse population of prompt hypotheses. (2) Evolutionary Reasoning Loop: This core phase iteratively refines the prompts. It employs a Visual Arena for pairwise competition to select elite individuals and uses Semantic Mutation to generate better prompt offspring, ensuring the semantic alignment of the generated masks. (3) Final Arbitration: A safeguard mechanism that leverages a double-blind swap strategy to arbitrate between the evolved text-based mask and a geometric intuition-based mask, ensuring robust segmentation results.}
    \label{fig:pipeline}
\end{figure*}

\section{Method}
\subsection{Problem Formulation}

We formally define the reasoning segmentation task as follows: given an input image $I \in \mathbb{R}^{H \times W \times 3}$ and a natural language query $q$, the objective is to generate a binary mask $M \in \{0, 1\}^{H \times W}$ that semantically aligns optimally with the target described by $q$. Traditional approaches, such as fully supervised fine-tuning or reinforcement learning, typically aim to learn a parameterized conditional distribution $P_\theta(M | I, q)$. However, this paradigm is inherently static during inference, relying heavily on the model's memorization of the training distribution.

In this work, we reformulate this task as an inference-time latent variable optimization problem. We posit that for any arbitrarily complex query $q$, there exists an optimal latent prompt $z^*$ within a hybrid semantic-geometric space $\mathcal{Z}$. When this optimal prompt drives a powerful, deterministic segmentation executor, it yields the optimal solution. Based on this premise, we define our objective as searching for $z^*$ within the space of frozen foundation models:

\begin{equation}
    z^* = \mathop{\arg\max}_{z \in \mathcal{Z}} \mathcal{F}(M_z, q; \Psi_{\text{VLM}}) \quad \text{s.t.} \quad M_z = \Phi_{\text{SAM}}(I, z)
    \label{eq:optimization_objective}
\end{equation}

In this formulation, $\Phi_{\text{SAM}}: \mathbb{R}^{H \times W \times 3} \times \mathcal{Z} \to \{0, 1\}^{H \times W}$ denotes the frozen Executor (implemented via SAM 3~\cite{sam3}), which deterministically maps the latent variable $z$ to a segmentation mask $M_z$. $\mathcal{F}$ represents the Fitness Function, parameterized by a frozen Vision-Language Model $\Psi_{\text{VLM}}$ (implemented via Qwen2.5-VL~\cite{Qwen2.5-VL}), which quantifies the semantic alignment between $M_z$ and the original query $q$. The core advantage of this modeling lies in decoupling: it leverages the MLLM's reasoning capabilities to navigate the complex semantic space $\mathcal{Z}$, while retaining SAM 3's robust geometric priors for pixel-level generation. Crucially, this avoids the risk of catastrophic forgetting associated with fine-tuning foundation models.

Solving the optimization problem in Eq.~\ref{eq:optimization_objective} presents two intrinsic mathematical challenges: non-differentiability and reference-free evaluation. First, since the search space $\mathcal{Z}$ consists of discrete natural language tokens and geometric coordinates, and $\Phi_{\text{SAM}}$ operates as a black-box function, the gradient $\nabla_z \mathcal{F}$ is inaccessible, precluding gradient-based optimization methods. Second, the absence of ground truth masks during inference makes the precise calculation of $\mathcal{F}$ impossible.

To address these challenges, we propose the EVOL-SAM3 framework, which employs an Evolutionary Algorithm (EA) as a derivative-free solver. Specifically, we maintain a population of hypotheses $\mathcal{P}_t = \{z_1^{(t)}, \dots, z_N^{(t)}\}$ at generation $t$, and approximate $z^*$ by iteratively applying MLLM-driven Semantic Mutation and Selection Operators. To overcome the lack of ground truth, we exploit the property that MLLMs exhibit superior performance in discrimination tasks compared to generation tasks. We approximate the fitness landscape via a Visual Arena mechanism, where the scalar fitness $\mathcal{F}(z)$ is modeled as the win-rate of a candidate solution in pairwise tournaments judged by $\Psi_{\text{VLM}}$. In this way, we transform static pre-trained models into a dynamic system capable of deliberate search and optimization.

\subsection{Framework Overview}

As illustrated in Figure~\ref{fig:pipeline}, EVOL-SAM3 reformulates the static reasoning segmentation task into a training-free, dynamic evolutionary search process. The framework leverages a frozen MLLM for high-level semantic planning and evaluation, while utilizing SAM 3 for low-level pixel execution. The pipeline initiates with Population Initialization (Phase 1), where the MLLM acts as a meta-generator to expand the user's single query into a diverse population of prompts covering various descriptive perspectives (e.g., attributes, spatial relations). Subsequently, the system enters the core Evolutionary Reasoning Loop (Phase 2). In each generation, candidate prompts drive SAM 3 to generate segmentation masks. These masks are rigorously filtered by a Semantic Gate for consistency and then evaluated via pairwise competition in a Visual Arena. Following the principle of survival of the fittest, elite individuals undergo MLLM-driven Semantic Mutation to spawn the next generation, iteratively converging towards the optimal semantic description of the target.

To mitigate potential linguistic hallucinations in complex scenarios, we introduce a final Heterogeneous Arena (Phase 3). In this stage, the system extracts not only the best text-derived mask from the evolutionary search but also generates an auxiliary detection-based mask derived from the MLLM's geometric intuition. These two candidates, originating from distinct reasoning paths (semantic reasoning vs. geometric localization), are submitted to a Heterogeneous Arena. Through a double-blind adaptive judgment mechanism, the MLLM selects the most robust final segmentation. This cascaded "Generate-Evolve-Arbitrate" mechanism enables Evo-SAM3 to achieve precise alignment and robust segmentation for complex visual queries without requiring any parameter updates.

\subsection{Semantic Meta-Planning}

The ultimate performance of evolutionary search is critically contingent on the quality and coverage of the initial population $\mathcal{P}_0$. In complex visual scenarios, the raw user query $q$ is often concise or semantically ambiguous (e.g., "the man" is ill-posed in a multi-person scene). Relying directly on such a single starting point makes the search process prone to getting trapped in local optima or failing to converge. To overcome this cold-start problem, we introduce a semantic meta-planning phase prior to the evolutionary loop. In this stage, the frozen MLLM functions as a vision-aware meta-generator. Unlike traditional blind textual augmentation or paraphrasing, this generator jointly analyzes the input image $I$ and query $q$ to deconstruct the abstract query into a set of visually grounded hypotheses. By identifying potential distractors within the image, it actively infers discriminative features capable of uniquely identifying the target, thereby extending a single query point into a distribution that covers the target's potential semantic space.

Specifically, the meta-planning process constructs a heterogeneous population $\mathcal{P}_0$ through multi-dimensional semantic anchoring. The first dimension is attribute augmentation, where the MLLM automatically supplements visual details (e.g., color, texture) to distinguish the target from similar instances. The second is spatial explicitness, which converts implicit relative positional relationships into absolute spatial descriptions (e.g., refining "the person on the left" to "the leftmost person" or "the person in the bottom-left corner"). The third is referent specification, which replaces generic pronouns with specific nouns. The initial population generated through this strategy possesses high genetic diversity and eliminates referential ambiguity to a significant extent at the initialization stage. This provides a robust and high-quality search space for the subsequent evolutionary reasoning loop, significantly reducing the exploration cost of the algorithm.

\begin{algorithm}[tb]
   \caption{Evolutionary Reasoning Loop}
   \label{alg:evo_loop}
   \begin{algorithmic}[1]
      \STATE {\bfseries Input:} Image $I$, Query $q$, Population $\mathcal{P}_0$, Max Gen $G$
      \STATE {\bfseries Output:} Optimal Mask $M^*$
      
      \FOR{$t=1$ {\bfseries to} $G$}
         \STATE $\mathcal{P}_{valid} \leftarrow \emptyset$
         
         \STATE \textit{\textbf{ Step A: Execution \& Cascaded Verification}}
         \FOR{$z_i \in \mathcal{P}_{t-1}$}
            \STATE $M_i \leftarrow \Phi_{\text{SAM}}(I, z_i)$
            \STATE $s_{conf} \leftarrow \text{Score}(M_i)$
            \STATE $g_{sem} \leftarrow \text{SemanticGate}(M_i, q; \Psi_{\text{VLM}})$
            \IF{$s_{conf} > \tau_{conf}$ \AND $g_{sem} == 1$}
               \STATE $\mathcal{P}_{valid} \leftarrow \mathcal{P}_{valid} \cup \{z_i\}$
            \ENDIF
         \ENDFOR

         \STATE \textit{\textbf{ Step B: Visual Arena Selection}}
         \STATE Initialize fitness $\mathcal{F}$ for all $z \in \mathcal{P}_{valid}$
         \FOR{pairs $(z_a, z_b)$ in $\mathcal{P}_{valid}$ with overlap}
            \STATE $w \leftarrow \text{VisualArena}(M_a, M_b, q; \Psi_{\text{VLM}})$
            \IF{$w = z_a$}
               \STATE Update $\mathcal{F}(z_a) \leftarrow \mathcal{F}(z_a) + \Delta$, \STATE $\mathcal{F}(z_b) \leftarrow \mathcal{F}(z_b) - \Delta$
            \ELSE
               \STATE Update $\mathcal{F}(z_b) \leftarrow \mathcal{F}(z_b) + \beta \Delta$, \STATE $\mathcal{F}(z_a) \leftarrow \mathcal{F}(z_a) - \beta\Delta$
            \ENDIF
         \ENDFOR
         
         \STATE $z_{elite} \leftarrow \arg\max_{z \in \mathcal{P}_{valid}} \mathcal{F}(z)$
         
         \STATE \textit{\textbf{Step C: Termination Check}}
         \IF{Converged($\mathcal{P}_{valid}$)}
            \STATE \textbf{break}
         \ENDIF
         
         \STATE \textit{\textbf{Step D: Semantic Mutation}}
         \STATE $\mathcal{P}_t \leftarrow \{z_{elite}\}$
         \STATE $z_{mut} \leftarrow \text{Mutation}(z_{elite}, q; \Psi_{\text{VLM}})$
         \STATE $\mathcal{P}_t \leftarrow \mathcal{P}_t \cup \{z_{mut}\}$
      \ENDFOR
      
      \STATE \textbf{return} $\Phi_{\text{SAM}}(I, z_{elite})$
   \end{algorithmic}
\end{algorithm}

\subsection{Detailed Mechanisms of Evolutionary Reasoning Loop}

To effectively optimize the latent prompt $z$ without ground truth supervision, we instantiate the evolutionary loop with three cascaded operators: Execution \& Cascaded Verification, Visual Arena Selection, and Semantic Mutation. This loop is designed to align discrete semantic search with continuous pixel generation, iteratively converging towards the optimal solution.

\paragraph{Execution and Cascaded Verification Operator.}
This step serves as the bridge between the discrete semantic space $\mathcal{Z}$ and the continuous pixel space, while simultaneously acting as a computational filter. For each latent prompt $z_i$ in the population $\mathcal{P}_t$, the frozen executor $\Phi_{\text{SAM}}$ deterministically maps it to a binary mask $M_i$ and an internal confidence score $s_{conf}^{(i)} \in [0, 1]$. This process is formulated as:
\begin{equation}
    (M_i, s_{conf}^{(i)}) = \Phi_{\text{SAM}}(I, z_i)
\end{equation}
To prune the search space of low-quality hypotheses, we employ a dual-layer gating function $\mathcal{G}(z_i)$. Initially, a \textit{Confidence Gate} leverages the executor's uncertainty to filter out hallucinations via an indicator function:
\begin{equation}
    g_{conf}(z_i) = \mathbb{I}(s_{conf}^{(i)} > \tau_{conf})
\end{equation}
Subsequently, candidates passing this preliminary filter are subjected to a \textit{Semantic Gate}, where the VLM $\Psi_{\text{VLM}}$ acts as a verifier. By analyzing the joint distribution of the raw image $I$ and the mask overlay $I \oplus M_i$, it validates consistency with the query $q$, yielding a binary decision $g_{sem}(z_i)$. Only individuals satisfying both constraints are admitted into the valid population:
\begin{equation}
    \mathcal{P}_{valid} = \{z_i \in \mathcal{P}_t \mid g_{conf}(z_i) \cdot g_{sem}(z_i) = 1\}
\end{equation}

\paragraph{Visual Arena Selection Operator.}
Approximating the fitness landscape $\mathcal{F}(z)$ in a reference-free setting is non-trivial. Exploiting the property that MLLMs are stronger discriminators than generators, we reformulate fitness estimation as a \textit{Pairwise Tournament}. For pairs $(z_a, z_b) \in \mathcal{P}_{valid}$ with significant geometric discrepancy, we construct a visual triplet input. The evaluator $\Psi_{\text{VLM}}$ functions as a judge, estimating the preference probability $P(z_a \succ z_b | q)$ based on semantic alignment. This relative ranking mechanism circumvents the calibration bias inherent in absolute scoring. The Visual Arena Selection Operator approximates the fitness landscape $\mathcal{F}(z)$ in a reference-free setting. Exploiting the property that MLLMs are stronger discriminators than generators, we reformulate fitness estimation as a Pairwise Tournament. For pairs $(z_a, z_b) \in \mathcal{P}_{valid}$ with significant geometric discrepancy, we construct a visual triplet input. The evaluator $\Psi_{\text{VLM}}$ functions as a judge, estimating the preference probability $P(z_a \succ z_b | q)$ based on semantic alignment. This relative ranking mechanism circumvents the calibration bias inherent in absolute scoring. We dynamically update individual fitness scores using an \textit{asymmetric} Elo-based update rule. To facilitate exploration, if the challenger $z_b$ defeats the incumbent $z_a$, we apply a boosted reward scaled by $\beta > 1$:
\begin{equation}
    \mathcal{F}(z_b) \leftarrow \mathcal{F}(z_b) + \beta \Delta, \quad \mathcal{F}(z_a) \leftarrow \mathcal{F}(z_a) - \beta \Delta
\end{equation}
Conversely, if the incumbent retains the advantage, a standard update step $\Delta$ is applied. This mechanism allows high-potential mutations to rapidly ascend in rank while maintaining population stability.

\paragraph{Semantic Mutation Operator.}
To escape local optima and traverse the semantic manifold effectively, we introduce a \textit{Semantic Mutation} operator $\mathcal{M}$ driven by the reasoning capabilities of the MLLM. Unlike stochastic character-level perturbations, this operator performs a logical reconstruction of the elite prompt $z_{elite}$ to generate a child $z_{mut}$ with higher "SAM-compatibility":
\begin{equation}
    z_{mut} = \mathcal{M}(z_{elite}, q; \Psi_{\text{VLM}})
\end{equation}
Our implementation incorporates two strategies to constrain the solution space of $\Phi_{\text{SAM}}$: (1) \textit{Simplification}, which removes non-discriminative redundancy to minimize executor attention drift; and (2) \textit{Ambiguity Avoidance}, which transforms relative spatial descriptions into absolute superlatives (e.g., mutating "person on the right" into "rightmost person"). These mutations explicitly guide the search towards regions of the semantic space that yield deterministic and accurate segmentation.

\subsection{Heterogeneous Arena}

While the evolutionary reasoning loop (Algorithm~\ref{alg:evo_loop}) excels in semantic alignment, producing a high-quality candidate $M_{text}$, relying solely on textual prompts remains susceptible to linguistic hallucinations in scenes with complex textures. To establish a robust safeguard, we introduce a heterogeneous arena module. This component leverages the geometric intuition of the MLLM to construct an auxiliary spatial reasoning path, facilitating a final adjudication against $M_{text}$ via a double-blind mechanism.

Geometric Intuition and Spatial Filtering. Distinct from the iterative refinement of textual prompts, the geometric path focuses on capturing the global spatial localization of the target via a single forward pass. We prompt the frozen MLLM to perform a detection task, directly predicting a bounding box $B_{pred}$, formulated as $B_{pred} = \Psi_{\text{VLM}}(I, q; \text{detect})$. This box serves as a geometric prompt for SAM to generate a candidate mask $M_{box} = \Phi_{\text{SAM}}(I, B_{pred})$. To ensure the validity of this geometric prior, we enforce a spatial consistency filter. This filter calculates the intersection over union between the bounding box of the generated mask, denoted as $\text{bbox}(M_{box})$, and the MLLM-predicted box $B_{pred}$. The geometric mask is deemed a valid contender only if it exhibits high spatial coherence:
\begin{equation}
    \text{IoU}(\text{bbox}(M_{box}), B_{pred}) > \tau_{spat}
\end{equation}
This constraint effectively discards instances where the MLLM localizes a general region but the segmentation model fails to identify a salient object within that vicinity, ensuring the geometric reliability of $M_{box}$.

Double-Blind Swap Mechanism. At this stage, we obtain two heterogeneous candidates: the optimal text-derived mask $M_{text}$ and the box-derived mask $M_{box}$. To select the final segmentation $M^*$ without bias, we submit them to a heterogeneous arena. Acknowledging the inherent position bias in MLLMs, where models may exhibit a preference for images appearing in specific input slots, we design a double-blind swap adjudication strategy. We construct two permuted visual prompt tuples: the forward permutation $V_{fwd} = (I, M_{text}, M_{box})$ and the reverse permutation $V_{rev} = (I, M_{box}, M_{text})$. The MLLM acts as a judge $J$ in two independent trials. The final selection is determined by a strict consistency logic that favors the robust geometric prior in uncertain cases:
\begin{equation}
    M^* = 
    \begin{cases} 
    M_{text} & \text{if } J(V_{fwd}) = M_{text} \land J(V_{rev}) = M_{text} \\
    M_{box} & \text{otherwise (Fallback Priority)}
    \end{cases}
\end{equation}
By enforcing consistency across permutations, we mitigate the stochasticity of the VLM. Specifically, any logical conflict (e.g., the model always choosing the first image regardless of content) or preference for the box triggers the fallback to $M_{box}$, thereby enforcing safe performance boundaries.

\section{Experiments}
\subsection{Implementation Details}
We implement EVOL-SAM3 using the PyTorch framework on a server equipped with 8 NVIDIA RTX 3080 GPUs. As a training-free framework, our method focuses on inference-time optimization without updating the backbone parameters. For the Multi-modal Large Language Model (MLLM) backbone, we employ Qwen2.5-VL-Instruct, evaluating both the 3B and 7B parameter versions to verify scalability. To optimize memory efficiency while maintaining performance, the MLLM inference is conducted in FP16 precision. For the visual segmentation backend, we utilize SAM 3, with image resolution and mask generation hyperparameters strictly aligned with the SAM3Agent baseline to ensure a fair comparison. The evolutionary process is configured with a population size of $N_{pop}=4$ and a maximum of $T_{max}=2$ generations. In the Visual Arena, the Elo rating update step is set to 1.0, and the confidence threshold for mask filtering is set to $\tau_{mask}=0.6$. All semantic mutations and arena selections are performed in a zero-shot manner.
\subsection{Datasets and Evaluation Metrics}
To ensure a fair comparison with state-of-the-art methods, we strictly follow the evaluation protocols and dataset settings established in previous research. We conduct extensive experiments on four standard benchmarks. For Reasoning Segmentation, we use the ReasonSeg dataset, which contains complex queries requiring implicit reasoning, and report results on its validation set. For Referring Expression Segmentation, we evaluate on RefCOCO, RefCOCO+, and RefCOCOg. Specifically, we use the \textit{val}, \textit{testA}, and \textit{testB} splits for RefCOCO and RefCOCO+; for RefCOCOg, we use the \textit{val (UMB)} and \textit{test (UMB)} partitions. Consistent with prior works, we employ gIoU and cIoU as the primary evaluation metrics.

\begin{table*}[t]
  \centering
  \caption{Comparison with state-of-the-art methods on ReasonSeg benchmarks for Reasoning Segmentation. Training-RES indicates whether the model has been fine-tuned on the Referring Expression Segmentation task. Training-ReasonSeg indicates whether the model has been fine-tuned on the ReasonSeg training set.}
  \label{tab:main_results}
  % 调整列间距
  \setlength{\tabcolsep}{8.5pt}
  \renewcommand{\arraystretch}{1.1}
  
  \begin{tabular}{l c cc cc cc cc cc}
    \toprule
    \multicolumn{2}{c}{Model} & \multicolumn{2}{c}{Training Data} & \multicolumn{2}{c}{Val Set} & \multicolumn{2}{c}{Test Set} & \multicolumn{2}{c}{Test (Short)} & \multicolumn{2}{c}{Test (Long)} \\
    \cmidrule(lr){1-2} \cmidrule(lr){3-4} \cmidrule(lr){5-6} \cmidrule(lr){7-8} \cmidrule(lr){9-10} \cmidrule(lr){11-12}
    Name & Version & RES & ReasonSeg & gIoU & cIoU & gIoU & cIoU & gIoU & cIoU & gIoU & cIoU \\
    \midrule
    SEEM~\cite{seem} & -- & $\times$ & $\times$ & 25.5 & 21.2 & 24.3 & 18.7 & 20.1 & 11.5 & 25.6 & 20.8 \\
    Grounded SAM~\cite{grounded_sam} & -- & $\times$ & $\times$ & 26.0 & 14.5 & 21.3 & 16.4 & 17.8 & 10.8 & 22.4 & 18.6 \\
    OVSeg~\cite{ovseg} & -- & $\times$ & $\times$ & 28.5 & 18.6 & 26.1 & 20.8 & 18.0 & 15.5 & 28.7 & 22.5 \\
    \midrule
    GLaMM~\cite{glamm} & Vicuna 7B & \checkmark & $\times$ & 47.4 & 47.2 & -- & -- & -- & -- & -- & -- \\
    SAM4MLLM~\cite{sam4mllm} & Qwen-VL 7B & \checkmark & $\times$ & 46.7 & 48.1 & -- & -- & -- & -- & -- & -- \\
    SAM4MLLM~\cite{sam4mllm} & LLaVA1.6 8B & \checkmark & $\times$ & 58.4 & 60.4 & -- & -- & -- & -- & -- & -- \\
    Seg-Zero~\cite{seg_zero} & Qwen2.5-VL 3B & \checkmark & $\times$ & 58.2 & 53.1 & 56.1 & 48.6 & -- & -- & -- & -- \\
    Seg-Zero~\cite{seg_zero} & Qwen2.5-VL 7B & \checkmark & $\times$ & 62.6 & 62.0 & 57.5 & 52.0 & -- & -- & -- & -- \\
    X-SAM~\cite{x_sam} & Phi3 3.8B & \checkmark & \checkmark & 56.6 & 32.9 & 57.8 & 41.0 & 47.7 & 48.1 & 56.0 & 40.8 \\
    HyperSeg~\cite{hyperseg} & Phi2 3B & \checkmark & \checkmark & 59.2 & 56.7 & -- & -- & -- & -- & -- & -- \\
    Ref-Diff~\cite{ref_diff} & LLaVA1.5 7B & $\times$ & $\times$ & -- & 52.4 & -- & 48.7 & -- & 48.0 & -- & 49.1 \\
    Ref-Diff~\cite{ref_diff} & LLaVA1.5 13B & $\times$ & $\times$ & -- & 60.5 & -- & 49.9 & -- & 48.7 & -- & 51.0 \\
    LISA~\cite{lisa} & LLaVA 7B & \checkmark & $\times$ & 44.4 & 46.0 & 36.8 & 34.1 & 37.6 & 34.4 & 36.6 & 34.7 \\
    LISA~\cite{lisa} & LLaVA 7B & \checkmark & \checkmark & 52.9 & 54.0 & 47.3 & 34.1 & 40.6 & 40.6 & 49.4 & 51.0 \\
    LISA~\cite{lisa} & LLaVA 13B & \checkmark & $\times$ & 48.9 & 46.9 & 44.8 & 45.8 & 39.9 & 43.3 & 46.4 & 46.5 \\
    LISA~\cite{lisa} & LLaVA 13B & \checkmark & \checkmark & 56.2 & 62.9 & 51.7 & 51.1 & 44.3 & 42.0 & 54.0 & 54.3 \\
    LISA~\cite{lisa} & Llama2 13B & \checkmark & \checkmark & 60.0 & 67.8 & 51.5 & 51.3 & 43.9 & 45.8 & 54.0 & 53.8 \\
    LISA~\cite{lisa} & LLaVA1.5 7B & \checkmark & $\times$ & 53.6 & 52.3 & 48.8 & 47.1 & 48.3 & 48.8 & 49.2 & 48.9 \\
    LISA~\cite{lisa} & LLaVA1.5 7B & \checkmark & \checkmark & 61.3 & 62.9 & 55.6 & 56.9 & 48.3 & 46.3 & 57.9 & 59.7 \\
    LISA~\cite{lisa} & LLaVA1.5 13B & $\times$ & \checkmark & 57.7 & 60.3 & 53.8 & 50.8 & 50.8 & 50.0 & 54.7 & 50.9 \\
    LISA~\cite{lisa} & LLaVA1.5 13B & \checkmark & \checkmark & 65.0 & \textbf{72.9} & 61.3 & 62.2 & 55.4 & 50.6 & 63.2 & 65.3 \\
    LENS~\cite{lens} & Qwen2.5-VL-3B & \checkmark & \checkmark & 62.1 & 64.9 & 57.2 & 58.0 & -- & -- & -- & -- \\
    HRSeg~\cite{hyperseg} & LLaVA1.1 7B & \checkmark & \checkmark & 57.4 & 54.7 & 43.4 & 41.7 & 54.4 & 54.5 & 51.6 & 50.5 \\
    HRSeg~\cite{hyperseg} & LLaVA1.6 7B & \checkmark & \checkmark & 64.9 & 63.1 & 49.5 & 48.7 & 59.2 & 58.9 & 57.0 & 57.2 \\
    SAM-Veteran~\cite{sam_veteran} & Qwen2.5-VL 7B & \checkmark & \checkmark & 68.2 & 67.3 & 62.6 & 56.1 & -- & -- & -- & -- \\
    SAM-Veteran~\cite{sam_veteran} & Qwen2.5-VL 32B & \checkmark & \checkmark & 72.3 & 70.0 & 62.9 & 58.2 & -- & -- & -- & -- \\
    READ~\cite{read_method} & LLaVA1.5 7B & \checkmark & \checkmark & 59.8 & 67.6 & 58.5 & 58.6 & 52.6 & 49.5 & 60.4 & 61.0 \\
    SegAgent~\cite{segagent} & Qwen2.5-VL 7B & \checkmark & \checkmark & 33.0 & 25.4 & 33.5 & 31.3 & -- & -- & -- & -- \\
    LLM-Seg~\cite{llm_seg} & LLaVA1.1 7B & \checkmark & \checkmark & 52.3 & 47.5 & 47.9 & 46.2 & 41.1 & 40.2 & 49.8 & 49.1 \\
    \midrule
    RSVP~\cite{rsvp} & LLaVA1.6 7B & $\times$ & $\times$ & 59.2 & 56.7 & 56.9 & 50.7 & 47.9 & 42.0 & 58.4 & 53.0 \\
    RSVP~\cite{rsvp} & Qwen2-VL 7B & $\times$ & $\times$ & 58.6 & 48.5 & 56.1 & 51.6 & 48.5 & 44.3 & 57.1 & 53.0 \\
    RSVP~\cite{rsvp} & Gemini1.5-Flash & $\times$ & $\times$ & 56.9 & 49.2 & 57.1 & 59.2 & 47.3 & 40.2 & 60.2 & 65.6 \\
    RSVP~\cite{rsvp} & GPT-4o & $\times$ & $\times$ & 64.7 & 63.1 & 60.3 & 60.0 & 55.4 & 50.4 & 61.9 & 62.5 \\
    Gemini Seg & Gemini2.5 Flash & ? & ? & 28.3 & 13.3 & 30.6 & 9.2 & 16.5 & 8.0 & 35.0 & 9.5 \\
    SAM 3 Agent~\cite{sam3} & Qwen2.5-VL 7B & $\times$ & $\times$ & 62.2 & 49.1 & 63.0 & 53.5 & 59.4 & 43.5 & 64.1 & 56.2 \\
    SAM 3 Agent~\cite{sam3} & Qwen2.5-VL 72B & $\times$ & $\times$ & \textbf{74.6} & 65.1 & 70.8 & 64.0 & \textbf{70.3} & 55.7 & 71.0 & 66.3 \\
    SAM 3 Agent~\cite{sam3} & Llama4 Maverick & $\times$ & $\times$ & 68.5 & 61.5 & 67.1 & 60.9 & 66.8 & \textbf{59.4} & 67.2 & 61.3 \\
    \midrule
    EVOL-SAM3 (\textbf{ours}) & Qwen2.5-VL 3B & $\times$ & $\times$ & 66.9 & 57.1 & 65.9 & 58.9 & 62.0 & 47.3 & 67.1 & 62.5 \\
    EVOL-SAM3 (\textbf{ours}) & Qwen2.5-VL 7B & $\times$ & $\times$ & 70.7 & 63.4 & \textbf{72.5} & \textbf{67.4} & 67.0 & 46.8 & \textbf{74.3} & \textbf{73.3} \\
    \bottomrule
  \end{tabular}
\end{table*}

\subsection{COMPARISONS}
As shown in Table \ref{tab:main_results}, we report detailed performance comparisons on the ReasonSeg benchmark. The most remarkable finding is that EVOL-SAM3 demonstrates superior performance under a fully zero-shot setting, surpassing existing learning-based methods (encompassing SFT and RL). Specifically, our 7B version achieves 70.7 gIoU on the validation set. This not only significantly outperforms the classic LISA-7B (52.9 gIoU) but, more notably, surpasses LISA-13B (65.0 gIoU), which has undergone deep optimization via supervised fine-tuning and reinforcement learning. This result prompts us to revisit the heavy reliance of high-performance reasoning segmentation on expensive supervised data or reward feedback, and demonstrates the feasibility of mining the reasoning potential of frozen models as an efficient alternative. 

Furthermore, among training-free counterparts, EVOL-SAM3 proves that a refined evolutionary mechanism is more efficient than merely scaling up model parameters. Although the baseline SAM 3 Agent achieves 74.6 gIoU on the validation set by relying on a massive 72B parameter model, our method achieves a reversal on the more challenging Test Set (72.5 vs. 70.8 gIoU) using only 7B parameters. Finally, the robustness of EVOL-SAM3 in handling long and complex queries is particularly pronounced. Benefiting from the effective decomposition and error correction of complex instructions enabled by the semantic mutation and visual arena mechanisms, we achieve an outstanding 74.3 gIoU on the Test (Long) subset. This significantly outperforms RSVP (GPT-4o) at 61.9 gIoU and SAM 3 Agent (72B) at 71.0 gIoU, fully validating the unique advantages of dynamic evolutionary search in deep semantic understanding.
\begin{table*}[htbp]
  \centering
  \caption{Comparison with state-of-the-art methods on Referring Expression Segmentation (RES) benchmarks. \textbf{Bold} indicates the overall best performance, and \underline{underline} indicates the best zero-shot performance among models not trained on RES datasets.}
  \label{tab:res_benchmark}
  \small
  \setlength{\tabcolsep}{6.5pt}
  \renewcommand{\arraystretch}{1.1}
  
  \begin{tabular}{l c c ccc ccc ccc}
    \toprule
    \multicolumn{2}{c}{Model} & Training & \multicolumn{3}{c}{RefCOCO} & \multicolumn{3}{c}{RefCOCO+} & \multicolumn{3}{c}{RefCOCOg} \\
    \cmidrule(lr){1-2} \cmidrule(lr){3-3} \cmidrule(lr){4-6} \cmidrule(lr){7-9} \cmidrule(lr){10-12}
    Name & Version & RES & val & testA & testB & val & testA & testB & val (U) & test (U) & val (G) \\
    \midrule
    LISA~\cite{lisa} & LLaVA 7B & \checkmark & 74.9 & 79.1 & 72.3 & 65.1 & 70.8 & 58.1 & 67.9 & 70.6 & -- \\
    GSVA~\cite{gsva} & 13B & \checkmark & 79.2 & 81.7 & 77.1 & 70.3 & 73.8 & 63.6 & 75.7 & 77.0 & -- \\
    GLaMM~\cite{glamm} & Vicuna 7B & \checkmark & 79.5 & 83.2 & 76.9 & 72.6 & 78.7 & 64.6 & 74.2 & 74.9 & -- \\
    SAM4MLLM~\cite{sam4mllm} & LLaVA1.6 7B & \checkmark & 79.6 & 82.8 & 76.1 & 73.5 & 77.8 & 65.8 & 74.5 & 75.6 & -- \\
    SAM4MLLM~\cite{sam4mllm} & LLaVA1.6 8B & \checkmark & 79.8 & 82.7 & 74.7 & 74.6 & 80.0 & 67.2 & 75.5 & 76.4 & -- \\
    GLEE~\cite{glee} & Plus & \checkmark & 79.5 & -- & -- & 68.3 & -- & -- & 70.6 & -- & -- \\
    GLEE~\cite{glee} & Pro & \checkmark & 80.0 & -- & -- & 69.6 & -- & -- & 72.9 & -- & -- \\
    DETRIS~\cite{detris} & DETRIS-L & \checkmark & 81.0 & 81.9 & 79.0 & 75.2 & 78.6 & 70.2 & 74.6 & 75.3 & -- \\
    UniLSeg~\cite{unilseg} & UniLSeg-20 & \checkmark & 80.5 & 81.8 & 78.4 & 72.7 & 77.0 & 67.0 & 78.4 & 79.5 & -- \\
    UniLSeg~\cite{unilseg} & UniLSeg-100 & \checkmark & 81.7 & 83.2 & 79.9 & 73.2 & 78.3 & 68.2 & 79.3 & 80.5 & -- \\
    PSALM~\cite{psalm} & Phi1.5 1.3B & \checkmark & 83.6 & 84.7 & 81.6 & 72.9 & 75.5 & 70.1 & 73.8 & 74.4 & -- \\
    EVF-SAM~\cite{evf_sam} & RC & \checkmark & 82.1 & 83.7 & 80.0 & 75.2 & 78.3 & 70.1 & 76.8 & 77.4 & -- \\
    EVF-SAM~\cite{evf_sam} & Extra Data & \checkmark & 82.4 & 84.2 & 80.2 & 76.5 & 80.0 & 71.9 & 78.2 & 78.3 & -- \\
    RICE~\cite{rice} & Qwen2.5-7B & \checkmark & 83.5 & 85.3 & 81.7 & \textbf{79.4} & 82.8 & 75.4 & 79.8 & 80.4 & -- \\
    MLCD-seg~\cite{mlcd_seg} & Qwen2.5-7B & \checkmark & 83.6 & 85.3 & 81.5 & \textbf{79.4} & 82.9 & \textbf{75.6} & 79.7 & 80.5 & -- \\
    HyperSeg~\cite{hyperseg} & Phi2 2.7B & \checkmark & 84.8 & 85.7 & \textbf{83.4} & 79.0 & \textbf{83.5} & 75.2 & 79.4 & 78.9 & -- \\
    LENS~\cite{lens} & Qwen2.5-VL 3B & \checkmark & 84.2 & 85.3 & 81.0 & \textbf{79.4} & 82.8 & 74.3 & 81.2 & 81.0 & -- \\
    SAM-Veteran~\cite{sam_veteran} & Qwen2.5-VL 7B & \checkmark & -- & 80.8 & -- & -- & 76.6 & -- & -- & 73.4 & -- \\
    SAM-Veteran~\cite{sam_veteran} & Qwen2.5-VL 32B & \checkmark & -- & 80.4 & -- & -- & 77.4 & -- & -- & 73.4 & -- \\
    READ~\cite{read_method} & LLaVA1.5 7B & \checkmark & 78.1 & 80.2 & 73.2 & 68.4 & 73.7 & 60.4 & 70.1 & 71.4 & -- \\
    SegAgent~\cite{segagent} & Qwen-VL 7B & \checkmark & 79.2 & 81.4 & 75.7 & 71.5 & 76.7 & 65.4 & 74.8 & 74.6 & -- \\
    X-SAM~\cite{x_sam} & Phi3 3.8B & \checkmark & \textbf{85.1} & \textbf{87.1} & \textbf{83.4} & 78.0 & 81.0 & 74.4 & \textbf{83.8} & \textbf{83.9} & -- \\
    \midrule
    GL-CLIP~\cite{gl_clip} & ResNet-50 & $\times$ & 32.7 & 35.3 & 30.1 & 37.7 & 40.7 & 34.9 & 41.6 & 42.9 & 44.0 \\
    GL-CLIP~\cite{gl_clip} & ViT-B/32 & $\times$ & 32.9 & 34.9 & 30.1 & 38.4 & 42.1 & 32.7 & 42.0 & 42.0 & 42.7 \\
    CaR~\cite{car_clip} & ViT-B/16 & $\times$ & 33.6 & 35.4 & 30.5 & 34.2 & 36.0 & 31.0 & 36.7 & 36.6 & 36.6 \\
    Ref-Diff~\cite{ref_diff} & VAE & $\times$ & 37.2 & 38.4 & 37.2 & 37.3 & 40.5 & 33.0 & 44.0 & 44.5 & 44.3 \\
    TAS~\cite{tas} & ResNet-50 & $\times$ & 39.9 & 42.9 & 35.9 & 44.0 & 50.6 & 36.4 & 47.7 & 47.4 & 48.7 \\
    TAS~\cite{tas} & ViT-B/32 & $\times$ & 39.8 & 41.1 & 36.2 & 43.6 & 49.1 & 36.5 & 46.6 & 46.8 & 48.1 \\
    IteRPrimeE~\cite{iterprime} & -- & $\times$ & 40.2 & 46.5 & 33.9 & 44.2 & 51.6 & 35.3 & 46.0 & 45.1 & 45.8 \\
    Pseudo-RIS~\cite{pseudo_ris} & CRIS & $\times$ & 39.8 & 44.8 & 33.0 & 42.2 & 46.3 & 34.5 & 43.7 & 43.4 & 43.8 \\
    Pseudo-RIS~\cite{pseudo_ris} & ETRIS & $\times$ & 41.1 & 48.2 & 33.5 & 44.3 & 51.4 & 35.1 & 46.0 & 46.7 & 46.8 \\
    LGD+DINO~\cite{lgd_dino} & ViT-B/32 & $\times$ & 49.5 & 54.7 & 41.0 & 49.6 & 58.4 & 38.6 & 50.3 & 51.1 & 52.5 \\
    VLM-VG~\cite{vlm_vg} & ResNet-50 & $\times$ & 47.7 & 51.8 & 44.7 & 41.2 & 45.9 & 34.7 & 46.6 & 47.1 & -- \\
    VLM-VG~\cite{vlm_vg} & ResNet-101 & $\times$ & 49.9 & 53.1 & 46.7 & 42.7 & 47.3 & 36.2 & 48.0 & 48.5 & -- \\
    HybridGL~\cite{hybridgl} & ViT-B/32 & $\times$ & 49.5 & 53.4 & 45.2 & 43.4 & 49.1 & 37.2 & 51.3 & 51.6 & -- \\
    SAM 3 Agent~\cite{sam3} & Qwen2.5-VL 7B & $\times$ & 59.4 & 64.3 & 55.0 & 51.4 & 57.0 & 44.9 & 57.2 & 58.8 & 59.7 \\
    \cmidrule(lr){1-12}
    EVOL-SAM3 (\textbf{ours}) & Qwen2.5-VL 3B & $\times$ & 66.8 & 70.9 & 59.9 & 56.1 & 63.1 & 49.3 & 63.0 & 63.5 & 64.0 \\
    EVOL-SAM3 (\textbf{ours}) & Qwen2.5-VL 7B & $\times$ & \underline{68.7} & \underline{73.7} & \underline{64.4} & \underline{64.4} & \underline{67.8} & \underline{54.0} & \underline{64.7} & \underline{65.5} & \textbf{\underline{65.9}} \\
    \bottomrule
  \end{tabular}
\end{table*}

Table \ref{tab:res_benchmark} presents the evaluation results on the classic RefCOCO, RefCOCO+, and RefCOCOg datasets. We observe that task-specific supervised training still yields dominant performance gains for RES tasks. As shown, fully fine-tuned methods (e.g., X-SAM, RICE, MLCD-seg) generally outperform zero-shot approaches by a large margin. This is attributed to the fact that RES tasks heavily rely on specific referring patterns (e.g., ``the person on the left''), which are explicitly reinforced during supervised training, enabling SFT models to establish strong text-region mappings. Consequently, zero-shot methods, lacking this targeted pattern adaptation, naturally face a performance gap.

However, within the Zero-Shot category, EVOL-SAM3 achieves a remarkable performance breakthrough, significantly narrowing the gap with supervised methods. Compared to existing training-free baselines, EVOL-SAM3 demonstrates superior generalization. Specifically, our 7B model achieves the best zero-shot performance across all three datasets. Compared to the baseline SAM 3 Agent, which shares the identical Qwen2.5-VL-7B backbone, our method realizes a comprehensive performance leap. For instance, on the RefCOCO \textit{val} and \textit{testA} subsets, we achieve improvements of +9.3 cIoU ($59.4 \to 68.7$) and +9.4 cIoU ($64.3 \to 73.7$), respectively. On RefCOCO+ \textit{val}, which demands strictly appearance-based grounding, the gain reaches +13.0 cIoU ($51.4 \to 64.4$). Similarly, on RefCOCOg \textit{val (U)}, which features more complex sentence structures, we obtain a significant increase of +7.5 cIoU ($57.2 \to 64.7$).

This significant performance improvement validates the necessity of transforming static inference into a dynamic iterative process. The results demonstrate that even in the absence of task-specific training data, introducing an adaptive optimization mechanism during the inference phase can effectively calibrate the misalignment between semantic understanding and visual perception, thereby achieving robustness and accuracy comparable to fully supervised methods in traditional referring expression segmentation tasks.

\subsection{ABLATION STUDY}
\subsubsection{Evolutionary Generations}
To thoroughly investigate the impact of evolutionary iterations on model performance, we conducted a sensitivity analysis on the maximum number of generations $T_{max}$ using the ReasonSeg validation set. As illustrated in the figure, the model exhibits a significant upward trajectory in segmentation performance as generations increase from 0 to 2, with gIoU and cIoU rising from 71.35\% and 66.30\% to 72.50\% and 67.39\%, respectively. This result strongly validates the effectiveness of our evolutionary mechanism: by introducing semantic mutation and arena selection, the model can transcend the limitations of the initial meta-planning and dynamically rectify early suboptimal solutions to approach more precise semantic descriptions. However, we observe distinct diminishing returns beyond $T_{max}=2$, where performance metrics saturate or even fluctuate slightly (e.g., a drop in cIoU at Gen 3). Given that the computational cost of inference increases linearly with evolutionary generations, further iterations do not yield performance gains commensurate with the additional overhead. Consequently, to achieve the optimal trade-off between reasoning accuracy and system efficiency, we adopt $T_{max}=2$ as the default configuration, which is sufficient for the model to achieve adequate reasoning calibration at a manageable cost.

We attribute the performance saturation and fluctuations observed beyond $T_{max}=2$ to two primary factors. First, the rapid convergence of the semantic search space. For most visual reasoning tasks, the number of valid logical paths from an ambiguous query to a precise referral is finite. Our experiments suggest that within two rounds of ``mutation-selection'' iteration, the evolutionary algorithm typically traverses and locks onto the most plausible semantic interpretation within the current context. Subsequent generations often involve minor paraphrasing rather than discovering substantive new semantic information. Second, the inherent upper bound of frozen backbones. It is crucial to note that EVOL-SAM3 is fundamentally a \textit{mining} strategy rather than a \textit{learning} process. It aims to elicit the maximum latent potential of the frozen MLLM and SAM but cannot surpass their intrinsic perceptual limits (e.g., recognizing extremely small or heavily occluded objects). Consequently, once the inference instruction is optimized to the boundary of the model's understanding, further evolutionary attempts risk introducing overly complex descriptions or hallucinations, leading to semantic drift. This also explains the slight oscillation in cIoU observed at higher generations.
\begin{figure}[htp]
    \centering
    \includegraphics[width=1\linewidth]{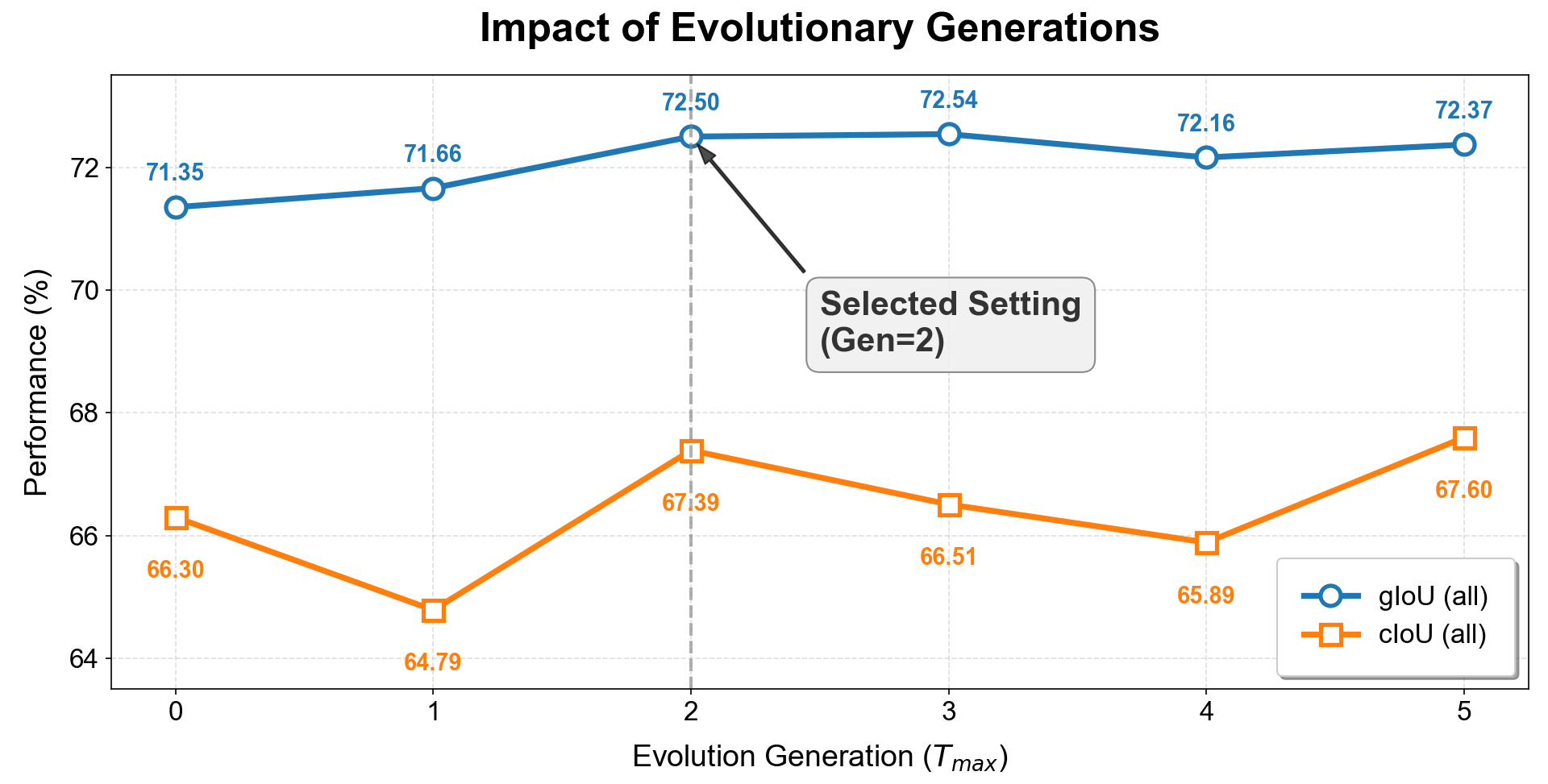}
    \caption{Ablation study on the number of evolutionary generations ($T_{max}$).}
    \label{fig:placeholder}
\end{figure}

\begin{table}[htp]
  \centering
  \small
  % 增加行高，使表格不拥挤
  \renewcommand{\arraystretch}{1.2}
  % 设置列宽，保证视觉平衡
  \setlength{\tabcolsep}{12pt}
  
  \caption{\textbf{Ablation study on the Visual Arena mechanism.} We compare our proposed Double-Switch comparative evaluation with Direct Scoring and Single-Pass strategies on the ReasonSeg validation set.}
  \label{tab:ablation_arena}
  
  \begin{tabular}{l cc}
    \toprule
    \textbf{Fitness Evaluation Strategy} & \textbf{gIoU} & \textbf{cIoU} \\
    \midrule
    Direct Scoring (w/o Arena) & 71.00 & 64.45 \\
    Single-Pass Judge (w/o Double-Switch) & 69.42 & 66.50 \\
    \midrule
    \textbf{Dual-Pass Arena (Ours)} & \textbf{72.50} & \textbf{67.40} \\
    \bottomrule
  \end{tabular}
\end{table}

\subsubsection{Effectiveness of the Visual Arena Mechanism}

Table~\ref{tab:ablation_arena} presents the ablation results regarding different fitness evaluation strategies. First, the Direct Scoring strategy, which requires the MLLM to assign absolute quality scores without peer comparison, yields a suboptimal cIoU of 64.45\%. This suggests that without a reference target, the model struggles to provide well-calibrated scores and is susceptible to numerical hallucinations, leading to unstable evolutionary directions. Second, introducing a Single-Pass Judge (pairwise comparison without position swapping) improves the cIoU to 66.50\% but, surprisingly, degrades the gIoU to 69.42\%. We attribute this drop to the inherent position bias of Large Language Models, where the model preferentially selects options at specific positions (e.g., typically favoring the first option). This bias results in the erroneous elimination of high-quality individuals during the selection phase. Finally, our Dual-Pass Arena, equipped with the Double-Switch mechanism, effectively mitigates this bias by enforcing consistency checks across swapped positions. This strategy achieves the best overall performance (72.50\% gIoU and 67.40\% cIoU), demonstrating that establishing a fair and robust comparison mechanism is critical for identifying superior solutions in evolutionary search.

\subsection{Qualitative Analysis}
\begin{figure*}[t]
    \centering
    \includegraphics[width=0.95\linewidth]{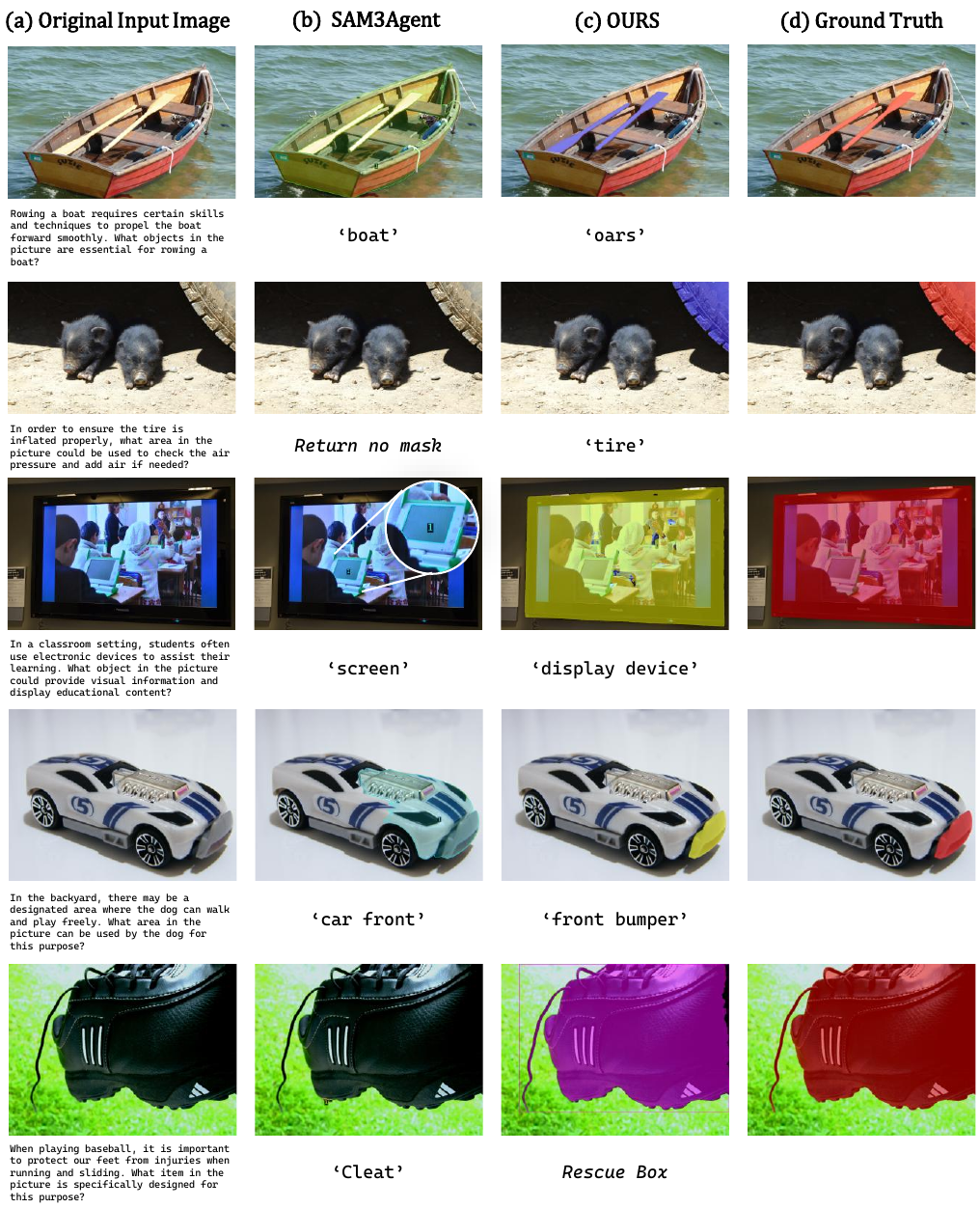}
    \caption{Qualitative comparison on the ReasonSeg benchmark.}
    \label{fig:visual}
\end{figure*}

To further evaluate the reasoning capability of the model in complex real-world scenarios, Figure \ref{fig:visual} presents a detailed visual comparison between EVOL-SAM3 and the strong baseline SAM3Agent. The results reveal the inherent limitations of static reasoning models in handling functional descriptions and abstract spatial referents, while our method demonstrates superior self-correction capabilities. First, the baseline model exhibits a strong salient object bias. When the query involves ``objects to propel the boat,'' the baseline captures the semantic relevance but is misled by the visually dominant ``boat,'' failing to localize the semantically correct but smaller ``oars.'' In contrast, EVOL-SAM3 leverages the semantic mutation mechanism within the evolutionary loop to successfully suppress this visual bias, precisely focusing attention on the specific components that align with the functional description.

Second, when distinguishing low-saliency or implicit targets, the baseline model is highly prone to linguistic hallucinations or missed detections. Benefiting from the incorporation of geometric intuition by the heterogeneous arena module, EVOL-SAM3 not only successfully recalls the imperceptible tire valve area but also accurately grounds the abstract concept of ``activity area'' to the specific grass field. This provides compelling evidence that modeling the reasoning process as a dynamic search rather than a single-pass mapping effectively bridges the gap between complex semantics and visual signals.

\section{Conclusion}
In this paper, we proposed EVOL-SAM3, a novel zero-shot framework that reformulates reasoning segmentation as an inference-time evolutionary search process. By synergizing a frozen MLLM with SAM 3 through a dynamic ``Generate-Evolve-Arbitrate'' pipeline, our method effectively navigates the semantic space to identify optimal prompts without any parameter updates. The introduction of the Visual Arena and Heterogeneous Arena mechanisms ensures robust optimization and safeguards against the hallucinations inherent in static inference. Extensive experiments on ReasonSeg and RefCOCO benchmarks demonstrate that EVOL-SAM3 not only surpasses existing training-free agents but also outperforms fully supervised state-of-the-art models in a zero-shot setting. This work provides compelling evidence that scaling inference-time computation via evolutionary strategies offers a promising and efficient alternative to traditional training paradigms for complex visual reasoning tasks.

\ifCLASSOPTIONcaptionsoff
  \newpage
\fi



\bibliographystyle{IEEEtran}
\bibliography{bibtex/bib/IEEEabrv,bibtex/bib/ref}

@article{Qwen2.5-VL,
  title={Qwen2.5-VL Technical Report},
  author={Bai, Shuai and Chen, Keqin and Liu, Xuejing and Wang, Jialin and Ge, Wenbin and Song, Sibo and Dang, Kai and Wang, Peng and Wang, Shijie and Tang, Jun and Zhong, Humen and Zhu, Yuanzhi and Yang, Mingkun and Li, Zhaohai and Wan, Jianqiang and Wang, Pengfei and Ding, Wei and Fu, Zheren and Xu, Yiheng and Ye, Jiabo and Zhang, Xi and Xie, Tianbao and Cheng, Zesen and Zhang, Hang and Yang, Zhibo and Xu, Haiyang and Lin, Junyang},
  journal={arXiv preprint arXiv:2502.13923},
  year={2025}
}

@article{sam3,
  title={{SAM} 3: Segment Anything with Concepts},
  author={Carion, Nicolas and others},
  journal={arXiv preprint arXiv:2511.16719},
  year={2025}
}

@inproceedings{seem,
  title={Segment Everything Everywhere All at Once},
  author={Zou, Xueyan and Yang, Jianwei and Zhang, Hao and Li, Feng and Li, Linjie and Gao, Jianfeng and Lee, Yong Jae},
  booktitle={Advances in Neural Information Processing Systems (NeurIPS)},
  year={2023}
}

@article{grounded_sam,
  title={{Grounded SAM}: Assembling Open-World Models for Diverse Visual Tasks},
  author={Ren, Tianhe and others},
  journal={arXiv preprint arXiv:2401.14159},
  year={2024}
}

@inproceedings{ovseg,
  title={Open-Vocabulary Semantic Segmentation with Mask-adapted {CLIP}},
  author={Liang, Feng and Wu, Bichen and Dai, Xiaoliang and Li, Kunpeng and Zhao, Yiming and Zhang, Hang and Zhang, Peizhao and Vajda, Peter and Marculescu, Diana},
  booktitle={Proceedings of the IEEE/CVF Conference on Computer Vision and Pattern Recognition (CVPR)},
  pages={7061--7070},
  year={2023}
}

@inproceedings{glamm,
  title={{GLaMM}: Pixel Grounding Large Multimodal Model},
  author={Rasheed, Hanoona and Maaz, Muhammad and Shaji, Sahal and Shaker, Abdelrahman and Khan, Salman and Cholakkal, Hisham and Anwer, Rao M and Xing, Eric and Yang, Ming-Hsuan and Khan, Fahad S},
  booktitle={Proceedings of the IEEE/CVF Conference on Computer Vision and Pattern Recognition (CVPR)},
  pages={13009--13018},
  year={2024}
}

@inproceedings{sam4mllm,
  title={{SAM4MLLM}: Enhance Multi-Modal Large Language Model for Referring Expression Segmentation},
  author={Chen, Yi-Chia and Li, Wei-Hua and Sun, Cheng and Wang, Yu-Chiang Frank and Chen, Chu-Song},
  booktitle={European Conference on Computer Vision (ECCV)},
  pages={323--340},
  year={2024}
}

@article{seg_zero,
  title={{Seg-Zero}: Reasoning-Chain Guided Segmentation via Cognitive Reinforcement},
  author={Liu, Yuqi and Peng, Bohao and Zhong, Zhisheng and Yue, Zihao and Lu, Fanbin and Yu, Bei and Jia, Jiaya},
  journal={arXiv preprint arXiv:2503.06520},
  year={2025}
}

@article{x_sam,
  title={{X-SAM}: From Segment Anything to Any Segmentation},
  author={Wang, Hao and Qiao, Limeng and Jie, Zequn and Huang, Zhijian and Feng, Chengjian and Zheng, Qingfang and Ma, Lin and Lan, Xiangyuan and Liang, Xiaodan},
  journal={arXiv preprint arXiv:2508.04655},
  year={2025}
}

@article{hyperseg,
  title={{HyperSeg}: Towards Universal Visual Segmentation with Large Language Model},
  author={Wei, Cong and Zhong, Yujie and Tan, Haoxian and Liu, Yong and Zhao, Zheng and Hu, Jie and Yang, Yujiu},
  journal={arXiv preprint arXiv:2411.17606},
  year={2024}
}

@inproceedings{lisa,
  title={{LISA}: Reasoning Segmentation via Large Language Model},
  author={Lai, Xin and Tian, Zhuotao and Chen, Yukang and Li, Yanwei and Yuan, Yuhui and Liu, Shu and Jia, Jiaya},
  booktitle={Proceedings of the IEEE/CVF Conference on Computer Vision and Pattern Recognition (CVPR)},
  pages={9579--9589},
  year={2024}
}

@article{rsvp,
  title={{RSVP}: Reasoning Segmentation via Visual Prompting and Multi-modal Chain-of-Thought},
  author={Lu, Yi and Cao, Jiawang and Wu, Yongliang and Li, Bozheng and Tang, Licheng and Ji, Yangguang and Wu, Chong and Wu, Jay and Zhu, Wenbo},
  journal={arXiv preprint arXiv:2506.04277},
  year={2025}
}

@inproceedings{gsva,
  title={{GSVA}: Generalized Segmentation via Multimodal Large Language Models},
  author={Xia, Zhuofan and Han, Dongchen and Han, Yizeng and Pan, Xuran and Song, Shiji and Huang, Gao},
  booktitle={Proceedings of the IEEE/CVF Conference on Computer Vision and Pattern Recognition (CVPR)},
  pages={3858--3869},
  year={2024}
}

@inproceedings{glee,
  title={{GLEE}: General Object Foundation Model for Images and Videos at Scale},
  author={Wu, Junfeng and Jiang, Yi and Liu, Qihao and Yuan, Zehuan and Bai, Xiang and Bai, Song},
  booktitle={Proceedings of the IEEE/CVF Conference on Computer Vision and Pattern Recognition (CVPR)},
  pages={3783--3795},
  year={2024}
}

@inproceedings{detris,
  title={{DETRIS}: Densely Connected Parameter-Efficient Tuning for Referring Image Segmentation},
  author={Huang, Jiaqi and Xu, Zunnan and Liu, Ting and Liu, Yong and Han, Haonan and Yuan, Kehong and Li, Xiu},
  booktitle={Proceedings of the AAAI Conference on Artificial Intelligence},
  pages={3653--3661},
  year={2025}
}

@inproceedings{unilseg,
  title={Universal Segmentation at Arbitrary Granularity with Language Instruction},
  author={Liu, Yong and Zhang, Cairong and Wang, Yitong and Wang, Jiahao and Yang, Yujiu and Tang, Yansong},
  booktitle={Proceedings of the IEEE/CVF Conference on Computer Vision and Pattern Recognition (CVPR)},
  pages={3459--3469},
  year={2024}
}

@inproceedings{psalm,
  title={{PSALM}: Pixelwise Segmentation with Large Multi-Modal Model},
  author={Zhang, Zheng and Ma, Yeyao and Zhang, Enming and Bai, Xiang},
  booktitle={European Conference on Computer Vision (ECCV)},
  pages={74--91},
  year={2024}
}

@article{evf_sam,
  title={{EVF-SAM}: Early Vision-Language Fusion for Text-Prompted Segment Anything Model},
  author={Zhang, Yuxuan and Cheng, Tianheng and Zhu, Lianghui and Hu, Rui and Liu, Lei and Liu, Heng and Ran, Longjin and Chen, Xiaoxin and Liu, Wenyu and Wang, Xinggang},
  journal={arXiv preprint arXiv:2406.20076},
  year={2024}
}

@article{rice,
  title={{RICE}: Region-Based Cluster Discrimination for Visual Representation Learning},
  author={Xie, Yin and Yang, Kaicheng and An, Xiang and Wu, Kun and Zhao, Yongle and Deng, Weimo and Ran, Zimin and Wang, Yumeng and Feng, Ziyong and Miles, Roy and others},
  journal={arXiv preprint arXiv:2507.20025},
  year={2025}
}

@inproceedings{mlcd_seg,
  title={Multi-label Cluster Discrimination for Visual Representation Learning},
  author={An, Xiang and Yang, Kaicheng and Dai, Xiangzi and Feng, Ziyong and Deng, Jiankang},
  booktitle={European Conference on Computer Vision (ECCV)},
  pages={428--444},
  year={2024}
}

@inproceedings{gl_clip,
  title={Zero-Shot Referring Image Segmentation with Global-Local Context Features},
  author={Yu, Seonghoon and Seo, Paul Hongsuck and Son, Jeany},
  booktitle={Proceedings of the IEEE/CVF Conference on Computer Vision and Pattern Recognition (CVPR)},
  pages={19456--19465},
  year={2023}
}

@inproceedings{car_clip,
  title={{CLIP} as {RNN}: Segment Countless Visual Concepts without Training Endeavor},
  author={Sun, Shuyang and Li, Runjia and Torr, Philip and Gu, Xiuye and Li, Siyang},
  booktitle={Proceedings of the IEEE/CVF Conference on Computer Vision and Pattern Recognition (CVPR)},
  pages={13171--13182},
  year={2024}
}

@article{ref_diff,
  title={{Ref-Diff}: Zero-Shot Referring Image Segmentation with Generative Models},
  author={Ni, Minheng and Zhang, Yabo and Feng, Kailai and Li, Xiaoming and Guo, Yiwen and Zuo, Wangmeng},
  journal={arXiv preprint arXiv:2308.16777},
  year={2023}
}

@article{tas,
  title={Text Augmented Spatial-Aware Zero-Shot Referring Image Segmentation},
  author={Suo, Yucheng and Zhu, Linchao and Yang, Yi},
  journal={arXiv preprint arXiv:2310.18049},
  year={2023}
}

@inproceedings{iterprime,
  title={{IteRPrimE}: Zero-Shot Referring Image Segmentation with Iterative Grad-CAM Refinement and Primary Word Emphasis},
  author={Wang, Yuji and Ni, Jingchen and Liu, Yong and Yuan, Chun and Tang, Yansong},
  booktitle={Proceedings of the AAAI Conference on Artificial Intelligence},
  pages={8159--8168},
  year={2025}
}

@inproceedings{pseudo_ris,
  title={{Pseudo-RIS}: Distinctive Pseudo-Supervision Generation for Referring Image Segmentation},
  author={Yu, Seonghoon and Seo, Paul Hongsuck and Son, Jeany},
  booktitle={European Conference on Computer Vision (ECCV)},
  pages={18--36},
  year={2024}
}

@article{lgd_dino,
  title={{LGD}: Leveraging Generative Descriptions for Zero-Shot Referring Image Segmentation},
  author={Li, Jiachen and Xie, Qing and Gu, Renshu and Xu, Jinyu and Liu, Yongjian and Yu, Xiaohan},
  journal={arXiv preprint arXiv:2504.14467},
  year={2025}
}

@inproceedings{vlm_vg,
  title={Learning Visual Grounding from Generative Vision and Language Model},
  author={Wang, Shijie and Kim, Dahun and Taalimi, Ali and Sun, Chen and Kuo, Weicheng},
  booktitle={IEEE/CVF Winter Conference on Applications of Computer Vision (WACV)},
  pages={8057--8067},
  year={2025}
}

@inproceedings{hybridgl,
  title={Hybrid Global-Local Representation with Augmented Spatial Guidance for Zero-Shot Referring Image Segmentation},
  author={Liu, Ting and Li, Siyuan},
  booktitle={Proceedings of the IEEE/CVF Conference on Computer Vision and Pattern Recognition (CVPR)},
  pages={29634--29643},
  year={2025}
}

@article{lens,
  title={{LENS}: Learning to Segment Anything with Unified Reinforced Reasoning},
  author={Zhu, Lianghui and Ouyang, Bin and others},
  journal={arXiv preprint arXiv:2508.14153},
  year={2025}
}

@inproceedings{mask_rcnn,
  title={Mask {R-CNN}},
  author={He, Kaiming and Gkioxari, Georgia and Doll{\'a}r, Piotr and Girshick, Ross},
  booktitle={Proceedings of the IEEE International Conference on Computer Vision (ICCV)},
  pages={2961--2969},
  year={2017}
}

@inproceedings{fcn,
  title={Fully Convolutional Networks for Semantic Segmentation},
  author={Long, Jonathan and Shelhamer, Evan and Darrell, Trevor},
  booktitle={Proceedings of the IEEE Conference on Computer Vision and Pattern Recognition (CVPR)},
  pages={3431--3440},
  year={2015}
}

@inproceedings{refcoco,
  title={{ReferItGame}: Referring Objects in Photographs of Natural Scenes},
  author={Kazemzadeh, Sahar and Ordonez, Vicente and Matten, Mark and Berg, Tamara},
  booktitle={Proceedings of the Conference on Empirical Methods in Natural Language Processing (EMNLP)},
  pages={787--798},
  year={2014}
}

@inproceedings{refcocog,
  title={Generation and Comprehension of Unambiguous Object Descriptions},
  author={Mao, Junhua and Huang, Jonathan and Toshev, Alexander and Camburu, Oana and Yu, Alan L and Murphy, Kevin},
  booktitle={Proceedings of the IEEE Conference on Computer Vision and Pattern Recognition (CVPR)},
  pages={11--20},
  year={2016}
}

@inproceedings{segagent,
  title={{SegAgent}: Exploring Pixel Understanding Capabilities in {MLLMs} by Imitating Human Annotator Trajectories},
  author={Zhu, Muzhi and Tian, Yuzhuo and Chen, Hao and Zhou, Chunluan and Guo, Qingpei and Liu, Yang and Yang, Ming and Shen, Chunhua},
  booktitle={Proceedings of the IEEE/CVF Conference on Computer Vision and Pattern Recognition (CVPR)},
  year={2025}
}

@inproceedings{read_method,
  title={{Reasoning to Attend}: Try to Understand How $<$SEG$>$ Token Works},
  author={Qian, Rui and Yin, Xin and Dou, Dejing},
  booktitle={Proceedings of the IEEE/CVF Conference on Computer Vision and Pattern Recognition (CVPR)},
  year={2025}
}

@inproceedings{llm_seg,
  title={{LLM-Seg}: Bridging Image Segmentation and Large Language Model Reasoning},
  author={Wang, Junchi and Ke, Lei},
  booktitle={Proceedings of the IEEE/CVF Conference on Computer Vision and Pattern Recognition Workshops (CVPRW)},
  pages={4689--4699},
  year={2024}
}

@article{sam_veteran,
  title={{SAM-Veteran}: An {MLLM}-Based Human-like {SAM} Agent for Reasoning Segmentation},
  author={Anonymous},
  journal={Under Review},
  year={2025},
  note={Cited as SAM-Veteran in benchmarks}
}
%\begin{thebibliography}{1}

%\end{thebibliography}



%\vfill


\end{document}